\documentclass[conference]{IEEEtran}
\IEEEoverridecommandlockouts
\usepackage{cite}
\usepackage{tabularx}
\usepackage{amsmath,amssymb,amsfonts}
\usepackage{algorithmic}
\usepackage{graphicx}
\usepackage{textcomp}
\usepackage{xcolor}
\usepackage[colorlinks=true,pdfstartview=FitV,linkcolor=black,citecolor=black,urlcolor=black,bookmarks=false]{hyperref}
\usepackage[caption=false]{subfig}
\def\BibTeX{{\rm B\kern-.05em{\sc i\kern-.025em b}\kern-.08em
    T\kern-.1667em\lower.7ex\hbox{E}\kern-.125emX}}
\begin{document}

\title{An Efficient and Scalable Deep Learning Approach for Road Damage Detection \\
}

\author{\IEEEauthorblockN{Sadra Naddaf-Sh\textsuperscript{1}, M-Mahdi Naddaf-Sh\textsuperscript{1}, Amir R. Kashani\textsuperscript{2}, Hassan Zargarzadeh\textsuperscript{1*}\thanks{*Corresponding author.}}
\IEEEauthorblockA{\textit{\textsuperscript{1}Phillip M. Drayer Electrical Engineering Department, Lamar University, Beaumont, Texas, USA} \\
\textit{\textsuperscript{2}Artificial Intelligence Lab, Stanley Oil \& Gas, Stanley Black \& Decker, Washington DC, USA}\\
% \textit{}\\
% Beaumont, Texas, USA \\
\{snaddafsharg, mnaddafsharg, hzargarzadeh*\}@lamar.edu, amir.kashani@sbdinc.com}
% \and
% \IEEEauthorblockN{2\textsuperscript{nd} M-Mahdi Naddaf-Sh}
% \IEEEauthorblockA{\textit{Phillip M. Drayer Electrical Engineering Department} \\
% \textit{Lamar University}\\
% Beaumont, Texas, USA \\
% mnaddafsharg@lamar.edu}
% \and
% \IEEEauthorblockN{3\textsuperscript{rd} Amir Kashani}
% \IEEEauthorblockA{\textit{Artificial Intelligence Lab} \\
% \textit{Stanley Oil & Gas, Stanley Black & Decker}\\
% MD, USA \\
% amir.kashani@sbdinc.com}
% \and
% \IEEEauthorblockN{4\textsuperscript{th} Hassan Zargarzadeh}
% \IEEEauthorblockA{\textit{Phillip M. Drayer Electrical Engineering Department} \\
% \textit{Lamar University}\\
% Beaumont, Texas, USA \\
% hzargarzadeh@lamar.edu}
% \and
% \and
}

% \IEEEoverridecommandlockouts
% \IEEEpubid{\makebox[\columnwidth]{978-1-7281-6251-5/20/\$31.00~\copyright2020 IEEE \hfill} \hspace{\columnsep}\makebox[\columnwidth]{ }}
\maketitle
% \IEEEpubidadjcol
 
\begin{abstract}
Pavement condition evaluation is essential to time the preventative or rehabilitative actions and control distress propagation. Failing to conduct timely evaluations can lead to severe structural and financial loss of the infrastructure and complete reconstructions. Automated computer-aided surveying measures can provide a database of road damage patterns and their locations. This database can be utilized for timely road repairs to gain the minimum cost of maintenance and the asphalt's maximum durability. This paper introduces a deep learning-based surveying scheme to analyze the image-based distress data in real-time. A database consisting of a diverse population of crack distress types such as longitudinal, transverse, and alligator cracks, photographed using mobile-device is used. Then, a family of efficient and scalable models that are tuned for pavement crack detection is trained, and various augmentation policies are explored. Proposed models, resulted in F1-scores, ranging from 52\% to 56\%, and average inference time from 178-10 images per second. Finally, the performance of the object detectors are examined, and error analysis is reported against various images. The source code is available at \hyperlink{https://github.com/mahdi65/roadDamageDetection2020}{https://github.com/mahdi65/roadDamageDetection2020}.

\end{abstract}


\section{Introduction} \label{Introduction}
Road infrastructure and its ability to efficiently and safely handle the transfer of people and goods point to point are vital means in societal, economic growth, and vitality. Long term viability of infrastructure and economy depend on strategies in managing, preserving, and rehabilitating of the road pavements. Moreover, poor pavement conditions are not only a contributing factor in excessive wear on vehicles, but also increase the number of crashes and delays that all lead to additional financial losses \cite{sun2016weighted}. The National Highway Traffic Safety Administration (NHTSA) claims that poor pavement conditions contribute to 16\% of traffic crashes \cite{national2008national}. 

Manual inspection is the dominant technique for pavement distress identification \cite{ouyang2010surface}. However, manual inspection can be labor-intensive, costly, and time-consuming. Furthermore, the manual inspection is prone to human visual error, safety issues because of the passing vehicles, and impeding the traffic flow \cite{gopalakrishnan2016advanced}. 

Performing maintenance operation on millions of miles of pavement is estimated to cost upwards of \$25 billion per year \cite{epps1997summary}. Pavement surveys, including surface and sub-surface assessments, are needed for effective and efficient maintenance operations. Frequent assessments are used in optimizing and prioritizing preservative and rehabilitative tasks. As current pavements age and new pavements are being added to the network, the manual inspection cannot satisfy timely assessment requirements, and it needs to be enhanced by semi-automated and ultimately automated methods. Moreover, automated pavement distress diagnosis techniques need to be accurate, cost-effective, non-destructive, generalizable, and relatively environmentally and user-friendly \cite{sollazzo2016hybrid, dargahi2020spatial}.

As a solution, deep learning-based methods show exceptional results in pavement distress detection and other applications in recent years \cite{naddaf2018design}. Zhang, et al. \cite{zhang2016road} utilized deep convolutional neural networks automatic pavement crack detection that was trained with manually annotated image patches acquired by a smartphone. Moreover, multiple approaches and datasets for pavement defect detection are reviewed in \cite{cao2020review} including multi-class crack classification \cite{naddaf2019real}, and segmentation \cite{jenkins2018deep}. By comparing multiple approaches of crack detection problem, Cao, et al. \cite{cao2020review} suggested further development in accuracy and real-time performance of the algorithms, robustness, and generalization in various weather conditions.

In a survey in \cite{cao2020survey}, eight models including Faster RCNN Resnet 50, Faster RCNN Resnet 101, Faster RCNN Inception Resnet, Faster RCNN Inception V2, SSD Mobile V1, SSD Mobile V2, SSD Inception V2, and SSDLite Mobile V2 are investigated for pavement crack detection. The authors reported that the highest mAP of 0.54 resulted from the Faster R-CNN Inception-Resnet-V2 model on a randomly selected validation set. Also, the inference time of the same model is estimated 17 seconds for a $600\times600$ pixel image, and 24.3 seconds for a $3680\times2760$ pixel image.

In addition, in object detection, there are some common methods like Ensemble Methods\cite{wbf2019,ensemblesbook} and Test-Time Augmentation(TTA) that are used to increase the performance of image recognition. These methods can significantly enhance the final prediction results. For example, \cite{rdd18wang} fused 16 models in the IEEE BigData Road Damage Detection Challenge 2018 and achieved a mean F1-score of 0.6455 that is the first place in the 2018 challenge. Additionally, top object detection competition winners with high accuracy detection applied similar methods e.g. \cite{gabruseva2020deep,Statoil/C-CORE}. However, the aforementioned method increases inference time, while being real-time is a favorable feature in the pavement crack detection task \cite{cao2020review}. 

There should be a shared database of cracks to justify and compare different method's performances and accuracy. Maeda, et al. \cite{maeda2018road} created an open-source crack dataset with annotations. This dataset is used for the IEEE BigData Cup Challenges of 2018, and 2020 \cite{arya2020}. This dataset contains annotated images from 3 countries, including India, Japan, and the Czech Republic. Since time performance is one of the challenges in the final crack detection platform \cite{cao2020review}, it is considered in this paper as one of the constraints. A family of EfficientDet models is trained and tested for evaluating the accuracy, robustness, and inference latency by utilizing the above road dataset. A series of trained models are prepared and tested utilizing the scalability feature of the selected models.  It is possible to select the model with respect to the target hardware without losing accuracy while the performance is maintained. A summery of competition results is provided in \cite{arya2020global}.

% So, the model selection, hyper-parameters, and training process are tuned such that a optimal balance between time performance and accuracy is achieved. 
% \textbf{ How? a paragraph is needed to be added below that explains.}

The remainder of this paper is organized as follows. Section \ref{Method} gives an overview of the applied method for pavement distress detection. In section \ref{Dataset}, the data preparation and dataset are explained. The model details, metrics, and experimental results are discussed in the section \ref{Experiments}. Finally, in the \ref{Conclusions} the overall results and future improvements are discussed. 

\section{Method} \label{Method}

In this paper, to address the trade-offs between accuracy, real-time performance, and scalability in pavement distress detection, a family of one-stage networks called EfficientDet \cite{effdet} is used. In contrast to one-stage methods, two-stage methods (like Faster-RCNN \cite{fasterrcnn}) can achieve higher accuracy by utilizing region proposal networks. On the other hand, scalability and real-time performance constraints make one-stage and one-scale designs a viable choice, maximizing the accuracy while minimizing the trained model's inference time. The EfficientDet is a single-stage, single-model, and single-scale object detector that can efficiently scale concerning the hardware resource constraints. Moreover, the trained model is deployable on end devices ranging from mobile-device with limited hardware and power resources to multi-GPU workstations. 

The first step would be feature extraction from a single image and then using object detection methods to locate cracks in the image. In EfficientDet the multi-scale feature extraction task is performed by EfficientNet backbone \cite{effnet}. In the pavement distress detection application, features like crack orientation, background, brightness, and the crack area should be taken into consideration. Hence, using multi-scale feature extraction can benefit from the accurate detection task. Feature Pyramid Network (FPN)\cite{fpn} uses a top-down approach to sum up multi-scale features that can be used for fusing multi-scale extracted features. In the FPN, different scales do not necessarily contribute equally to the output features that can lead to some missing features in the crack detection process. 

% In this paper, the same backbone is used for feature extraction; instead, the development is done on the feature selection that is applicable for crack detection application \textbf{this sentence is not clear}.
% (\textbf{you have not explained the layers yet})

Bi-directional Feature Pyramid Network (BiFPN) can be used to address the equal contribution issue in FPN. Also, BiFPN exploits trainable weights to learn features with the most contribution in the final model. So, features from layers P3 to P7 from the backbone are passed to BiFPN as the selected multi-scale features. Finally, to obtain the class and box of detected cracks in an image, the output of BiFPN will be fed into box and class prediction networks. 

\begin{figure*}[htbp]
\centering
\centerline{\includegraphics[scale=.20]{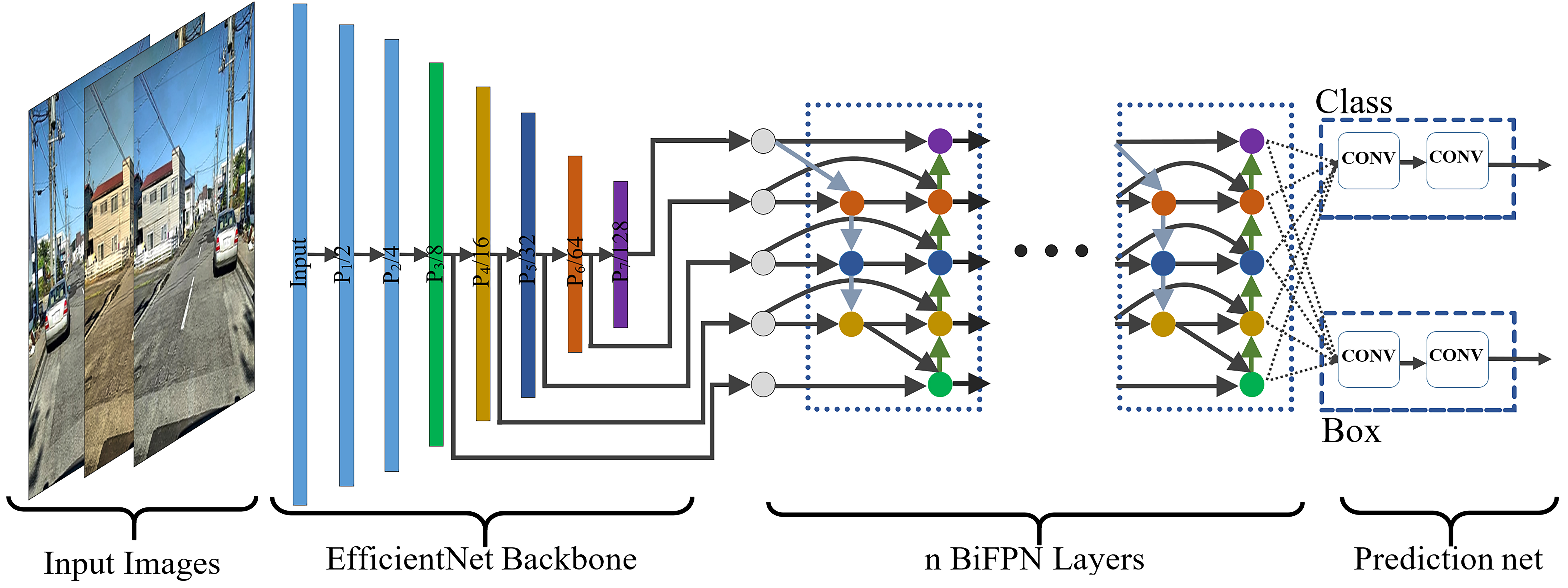}}
     \caption{EfficientDet family architecture: Images pass through the backbone, and feature scales P3 to P7 get fed into the BiFPN network. Input image resolution is calculated from \eqref{res}. Number of BiFPN layers extracted using \eqref{bifpn}. Depth of box/class prediction net is determined using \eqref{box/class}.}
  \label{figeffdet} 
\end{figure*}

The depth and width of EfficientDet's backbone are following the width and depth scaling coefficients of EfficientNet \cite{effnet}. The BiFPN and box/class prediction network depth and width in the EfficientDet are also calculated separately. Each model from EfficientDet-D0 to EfficientDet-D7 have a similar architecture to figure \ref{figeffdet}. It should be noted that from D0 to D7, the size of the backbone, resolution of the input image, the number of repeated BiFPN blocks, and the number of layers in box/class networks increases using a single compound coefficient $\phi$. Therefore, using equations \eqref{res}, \eqref{bifpn}, \eqref{box/class} input image resolution, BiFPN depth and width, depth of box/class networks are determined, respectively. Tables \ref{tabbigtable} and \ref{tableinf} represent the input image resolution and parameters devised for each model.
\begin{equation}
R\textsubscript{input}= 512 +\phi \cdot 128
\label{res}
\end{equation}
\begin{equation}
W\textsubscript{bifpn} = 64\cdot(1.35\textsuperscript{$\phi$}), D\textsubscript{bifpn} = 3 + \phi
\label{bifpn}
\end{equation}
\begin{equation}
D\textsubscript{box}=D\textsubscript{class}= 3 + \lfloor \frac{\phi}{3}\rfloor  
\label{box/class}
\end{equation}

\begin{table}[tbp]
\caption{Distribution of Images and Labels}
\begin{center}
\begin{tabular}{|c|c|c|c|c|c|}
\hline
\textbf{}& \textbf{Total} & \multicolumn{4}{|c|}{\textbf{Categories}} \\
\cline{3-6} 
\textbf{} & &\textbf{\textit{D00}}& \textbf{\textit{D10}}& \textbf{\textit{D20}}  & \textbf{\textit{D40}} \\
\hline
\textbf{Train}& 18930 &  5918 & 4014 & 7535 & 5103 \\
\hline
\textbf{Validation}& 2111 & 674 & 432 & 846 & 524 \\
\hline
\textbf{Test1}& 2631& \textbf{---}$^{\mathrm{a}}$&\textbf{---} &\textbf{---}&\textbf{---} \\
\hline
\textbf{Test2}& 2664&\textbf{---} &\textbf{---} &\textbf{---}&\textbf{---} \\
\hline
\textbf{Total}& 26336 & 6592 & 4446 & 8381 & 5627 \\
\hline
\multicolumn{4}{l}{$^{\mathrm{a}}$ Testset annotations were not released.}
\end{tabular}
\label{tabdataset}
\end{center}
\end{table}

\section{Dataset} \label{Dataset}

In this paper, the presented dataset in IEEE BigData 2020 Road Damage Detection challenge is used \cite{maeda2018road}. Images are taken using smartphones mounted on the dashboard vehicles. Annotations for this dataset are provided in PASCALVOC format\cite{pascalvoc}. Crack types are named based on the Japan Road Association(JRA)\cite{jra} standards. The cracked area is annotated using a rectangular box, and each box is associated with a crack class label. In the 2020 challenge, four classes of interest consist of Longitudinal crack(D00), Transverse crack(D10), Alligator Crack(D20), and Potholes(D40).

The distribution of classes in the dataset is presented in Table \ref{tabdataset}. A total of 26336 images are available in the dataset. Challenge organizers withheld 20\% of images (i.e., 5295) as test a set and released the rest of the images (i.e., 21041) with annotations to the contestants. The test set is divided into two sections; Test1 includes 2631 images, and Test2 includes 2664 images. The test set annotations are not released to the participants. The evaluation of the test set is performed online and with a limited number of tries per day, and the feedback is F1-score.

\section{Experiments} \label{Experiments}
In this section, firstly, metrics including F1-score and Average Precision(AP) for analyzing and evaluating the results, are discussed. Then, network training, augmentation transferring, and hyperparameter tuning for pavement crack detection are studied. Finally, in a discussion, multiple models trained with various parameters and their contribution to the accuracy, time performance, and F1-score are compared.

\subsection{Evaluating Metrics}

For a fair comparison of experiment results, two sets of metrics are applied. First, measuring the F1-Score due to competition rules and also to submit/compare results with other participating teams. Second, Average Precision is used to be able to measure precision-recall on different confidence thresholds. %used on MS COCO \cite{mscoco} as well as other state of the art object detection papers\cite{effdet,effnet,panet}.

\textbf{F1-Score:} Competition Defined metrics as used in \cite{arya2020} are as the following: 

\begin{equation}
F1 = 2 \cdot \frac{ precision \cdot recall}{precision+ recall}
\end{equation}
where, 

\begin{equation}
precision = \frac{tp}{tp + fp} ;  recall = \frac{tp}{tp + fn}   
\end{equation}
%A box is counted as correct/incorrect detection based on the area of its overlap and the area of ground truth box.
in each of which $tp$ is equal to the number of true positive detections of the model, $fp$ is the number of objects that model incorrectly marked as crack, and $fn$ indicates the number of existing objects(cracks) in the image that predictor has not detected. A bounding box is considered as correct when the area of the ground truth bounding box and the detected box have at least 0.5(based on competition rule) Intersection over the area of the union of two boxes, which is called Intersection over Union(IoU). Also, the class labels of both bounding boxes should be the same, that means:
\begin{equation}
IoU = \frac{area(B\textsubscript{Bp} \cap B\textsubscript{gt})}{area(B\textsubscript{Bp} \cup B\textsubscript{gt})}.  
\end{equation}
% IOU = area(Bp \ Bgt)
% area(Bp [ Bgt);
%f1 problems : might be biased toward the class with most tp/fp(for example in this case ... ). finding the best threshold every on model to maximize performance.   
% IoU is defined as below : 

\textbf{Average Precision (AP):} is also used to demonstrate and compare results. F1-score results are reported using only one confidence level. However, recall and precision of a robust object detector are not altering much, with varying confidence. The area under curve for precision-recall curve will not get precise results due to zig-zag like curve\cite{odm}. Instead, defining all-point interpolation can obtain accurate results by pruning zig-zag behavior of the precision-recall curve. AP has become a standard for comparing model performance in different object detection challenges\cite{cococomptition,mscoco,pascalvoc,vocpascal} as well as in literature\cite{yolo9000,effdet,fasterrcnn,retina}. In section \ref{evalres}, models are evaluated using mAP(mean AP)(where with setting IOU threshold from 0.50 to 0.95 with 0.05 step mAP is calculated), AP\textsubscript{50}, AP\textsubscript{75}, AP\textsubscript{s}(s stands for small and objects with area $<$ 32\textsuperscript{2}) , AP\textsubscript{m}(m is medium and area of the objects are between 32\textsuperscript{2} and 96\textsuperscript{2}), as well as AP\textsubscript{l}(objects with area $>$ 96\textsuperscript{2}).
% we used COCOAPI \cite{mscoco} for AP and \cite{odm} for F1-score.

\begin{table*}[htb]
\caption{Model Performance Details }
\begin{center}
\begin{tabular}{|c|cccc|ccccccc|}
\hline
% \textbf{}& \textbf{Total} & \multicolumn{4}{|c|}{\textbf{Categories}} \\
% \cline{3-6} 
\begin{tabular}{@{}c@{}}   \textbf{\textit{Model}} \\ \textbf{\textit{Name}}\end{tabular} &
\begin{tabular}{@{}c@{}}   \textbf{\textit{Input Image}} \\ \textbf{\textit{Resolution}}\end{tabular} &
\begin{tabular}{@{}c@{}}   \textbf{\textit{Backbone}} \\ \textbf{\textit{Name}}\end{tabular} &
% \begin{tabular}{@{}c@{}}   \textbf{\textit{BiFPN}} \\ \textbf{\textit{\#Channels}}\end{tabular} &
% \begin{tabular}{@{}c@{}}   \textbf{\textit{BiFPN}} \\ \textbf{\textit{\#layers}}\end{tabular} &
% \begin{tabular}{@{}c@{}}   \textbf{\textit{Box/Class}} \\ \textbf{\textit{\#layers}}\end{tabular} &
% \begin{tabular}{@{}c@{}}   \textbf{\textit{Model}} \\ \textbf{\textit{Size(MB)}}\end{tabular} &
%  \textbf{\textit{\#Params}} &

% \begin{tabular}{@{}c@{}}   \textbf{\textit{Anchor}} \\ \textbf{\textit{ratios}}\end{tabular} &
% \begin{tabular}{@{}c@{}}   \textbf{\textit{batch size/}} \\ \textbf{\textit{learning rate}}\end{tabular} &

% \begin{tabular}{@{}c@{}c@{}}   \textbf{\textit{Inference}} \\ \textbf{\textit{Time(ms)}} \\ V100\end{tabular} &
\begin{tabular}{@{}c@{}}   \textbf{\textit{Test1}} \\ \textbf{\textit{F1}}\end{tabular} &
\begin{tabular}{@{}c@{}}   \textbf{\textit{Test2}} \\ \textbf{\textit{F1}}\end{tabular} &

\multicolumn{7}{c|}{\textbf{\textit{Validation}}} \\
% \cline{8-10} 
&&&& &  \textbf{\textit{AP}}  \textbf{\textit{}}&
 \textbf{\textit{AP\textsubscript{50}}}  \textbf{\textit{}} &
   \textbf{\textit{AP\textsubscript{75}}}  \textbf{\textit{}} &
      \textbf{\textit{AP\textsubscript{s}}}  \textbf{\textit{}} &
   \textbf{\textit{AP\textsubscript{m}}}  \textbf{\textit{}} &
   \textbf{\textit{AP\textsubscript{l}}}  \textbf{\textit{}} &
   \textbf{\textit{F1}} \\% \textbf{\textit{}} \\

\hline
\textbf{D0}& 512 &  B0 &  52.1&51.4 & 19.1 & 47.2 &11.5 & 7.2 &14.3&22.2&54.04 \\
% \hline
\textbf{D0-AUG}& 512 &  B0 &  51.2&52.5 & 19.8 & 48.4 &12.1 & 7.9 &15.4 & 22.7&54.03 \\
\hline

\textbf{D1}& 640 & B1 &  53.8&54.7 & 21.7 & 51.5 & 13.4 & 15.3 & 16.9 & 25.0 & 56.9  \\
% \hline
\textbf{D1-AUG}& 640 & B1 &  54.4 &\textbf{55.4} & 22.0 & 51.7 & 13.1 & 17.1 & 17.7 & 24.7 & 56.5  \\
\hline

\textbf{D2}& 768& B2&  55.2 & 54.9 & 22.9 & 53.5 & 14.9&10.4& 18.6 &24.9 &56.7 \\
\textbf{D2-AUG}& 768& B2&  54.1 & 54.0 & 22.9 & 54.2 & 15.2&13.3& 18.8 &24.7 &56.6 \\

\hline
\textbf{D3}& 896&B3 &  \textbf{56.5} & 54.7 & 23.0 & 53.4 &15.0&10.5 &18.4& 25.4 &56.5\\
\textbf{D3-AUG}& 896&B3 &  56.3 & 54.2 & 22.6 & 53.4 &14.7&11.4 &18.3& 24.8 &56.8\\
\hline
\textbf{D4}& 1024 & B4 & 54.53&54.6 & 22.8 & 53.3 &15.1 &15.5& 18.1 &25.7&57.2 \\
% \textbf{D5}& 1280 & B5 & ?&? & ap & ap50 &? &?& ? &?&? \\
\hline
% \textbf{D6}& 1280 & B6 & ?&? & ap & ap50 &? &?& ? &?&? \\

\textbf{D7-AUG}& 1536 & B6 & \textbf{56.5}&\textbf{54.9} & 23.4 & 53.6 &15.0 &31.4& 19.2 &25.5&56.5 \\

% \textbf{D4}& 26336 & 6592 & 4446 & 8381 & 5627 \\

\hline
% \multicolumn{4}{l}{$^{\mathrm{*}}$ Test sets annotations were not released.}
\end{tabular}
\label{tabbigtable}
\end{center}
\end{table*}

\begin{table}[htbp]
\caption{Inference Time and other parameters related to each model.}
\begin{center}
\begin{tabular}{|c|c|c|ccc|}
\hline

\begin{tabular}{@{}c@{}}   \textbf{\textit{Model}} \\ \textbf{\textit{Name}}\end{tabular} &
% \begin{tabular}{@{}c@{}}   \textbf{\textit{Input Image}} \\ \textbf{\textit{Resolution}}\end{tabular} &
% \begin{tabular}{@{}c@{}}   \textbf{\textit{BackBone}} \\ \textbf{\textit{Name}}\end{tabular} &
% \begin{tabular}{@{}c@{}}   \textbf{\textit{Model}} \\ \textbf{\textit{Size(MB)}}\end{tabular} &
 \textbf{\textit{\#Params}} &
\begin{tabular}{@{}c@{}}   \textbf{\textit{Batch size/}} \\ \textbf{\textit{Learning rate}}\end{tabular}&
\multicolumn{3}{c|}{\begin{tabular}{@{}c@{}c@{}}  \textbf{\textit{Inference}}  \textbf{\textit{Time(img/s)}} \\ \textbf{\textit{V100}}\end{tabular}} \\
&&& b$^{\mathrm{b}}$=1& b$^{\mathrm{b}}$=8& b$^{\mathrm{b}}$=16 \\
\hline
\textbf{D0}&   3.8M & 90/0.112 & 20&121&178\\
\hline
\textbf{D1} & 6.5M & 75/0.075 & 15&98&147\\
\hline
% \textbf{D1-AA}   & ?& ?M & ?/? & ?  \\
% \hline
\textbf{D2}&  8M &45/0.056 & 12&82&100 \\
\hline
\textbf{D3} &  11.9M &18/0.026 & 10 &54&58\\
\hline
\textbf{D4}  &  20.5M & 9/0.011 & 9 &35&37    \\
\hline

\textbf{D7}  & 51M & 8/0.01 & 6 &9&10\\

\hline

\multicolumn{4}{l}{$^{\mathrm{}}$ b denotes batch size for inference.}
\end{tabular}
\label{tableinf}
\end{center}
\end{table}

\subsection{Anchor boxes}
EfficientDet box prediction network utilizes anchor boxes to enhance the detection of overlapping bounding boxes. The default value for the box prediction network consists of three aspect ratios (i.e. 0.5, 1.0, 2.0). For the crack detection task, based on all labeled bounding boxes in the dataset, the default value is increased to seven (and in larger models eight) aspect ratios. Also, k-means clustering \footnote{partially used https://github.com/mnslarcher/kmeans-anchors-ratios} is used to find the set of optimal aspect ratios for box prediction network \cite{yolo9000}. Moreover, the input image resolution is also considered in aspect ratio calculation. Other parameters left as it is in the code repository of \cite{effdet}.
% improved model performance (+3 AP\textsubscript{50}).

\subsection{Training and Hyperparameters Tuning:}
For training the EfficientDet D0-D4 models, a GPU cluster with three V100-16GB NVIDIA GPUs is used. The input image size in EfficientDet D7 is 2.5x of the original image size. Hence, more computational power is needed, and the model is trained on a GPU cluster with eight V100-32GB NVIDIA GPUs. Also, a technique for fitting large models in limited GPU memory is to employ Gradient Checkpointing \cite{grad}, even though training time increases significantly. Batch size is proportional to the model size due to GPU memory limitation. This limitation is more observable in D3-D7 models than D0-D2. In distributed training, synchronized cross-GPU batch normalization(syncBN) is applied, which provides access to cross-device batch normalization to improve statistics.

During training, before each image is fed to the network, it is first randomly flipped horizontally and/or resized and finally normalized using mean and standard deviation values that are over each RGB channel. Also, for all models, mixed-precision training is used. Mixed precision\cite{mixed} utilizes performing half-precision operations, leading to faster inference and training and decreased memory usage. In the training process, decreasing memory usage makes it possible to fit larger models and higher-resolution images to the GPUs.\footnote{apex package for mixed-precision training available at https://nvidia.github.io/apex} Moreover, pre-trained weights on MS COCO \cite{effdet} that are converted to Pytorch \cite{ptheffdet} are utilized as initial weights. Furthermore, all layers are trained, and the layers' weights are not frozen.

The learning rate and batch sizes are listed in table \ref{tableinf}. Also, the cosine learning rate is used \cite{loshchilov2016sgdr}. In the beginning of training for the first few epochs (3-5) learning rate increases gradually to the desired point, and from epoch 5 to the end of the training process learning rate decreases gradually in a cosine form. In addition, learning rate noises applied to 30\% and 90\% of the training process. Moreover, an exponential moving average with weight decay of 0.9998 is applied to stabilize training, especially on larger models that are trained with relatively small batch size. Non-Maximum Suppression(NMS) is applied to prune overlapping bounding boxes through inference. In addition, based on validation results, the best threshold per class is extracted and applied during inference. Table \ref{tableinf} illustrates the inference latency and model size for each model.

\subsection{Transferring Augmentation Policies}
Data augmentation not only improves accuracy but also generalizes the trained model \cite{obdataAug}. There are countless possibilities for augmentation strategy, including geometric transforms, brightness and color balance adjustment, histogram equalization, and rotation. It is possible to use multiple augmentation strategies on a single image with a varying range of magnitudes. However, selecting optimal augmentation strategies that improve the results and enhance performance is a time-consuming task. Some techniques are developed, such as AutoAugment \cite{autoaugment}, and RandAungmet \cite{randaugment} to address this issue, which can extract the best set of policies through reinforcement learning. For finding the optimal policy, the model should be trained and evaluated with each set of proposed policies. There are four default policies introduced in \cite{obdataAug}, which are called policyV0-policyV3.  One of the shared augmentation strategies in policyV0 to policyV3 is rotation. By using the rotation strategy, crack orientation will change, i.e., it converts a Longitudinal crack(D00) to Transverse crack(D10) or conversely. Therefore, the policies V0-V3 are updated by eliminating this strategy. 

A D0 model with randomly selected 25\% of the training set is trained for 150 epochs and evaluated on the validation set with all V0-V3 policies. For comparison, the same process is repeated without applying any augmentation. As it is depicted in Table \ref{tableaugpolicies}, policies V1 and V2 increased AP metrics more than others. Since all AP metrics are increased using policy V1, it is selected as the augmentation policy on the entire dataset during training. It should be noted that the augmentation strategies introduced in \cite{obdataAug} use bounding box-only augmentation. In this paper, the same augmentation strategies are applied to the entire image.

\begin{table}[tbp]
\caption{Policies used to train on 25\% of train images.}
\begin{center}
\begin{tabular}{|c|cccc|}
\hline

\begin{tabular}{@{}c@{}}   \textbf{\textit{policy}} \\ \textbf{\textit{Name}}\end{tabular} &
\multicolumn{4}{c|}{\textbf{\textit{Base ( \% of Improvement)}}} \\%& \multicolumn{4}{c|}{\textbf{\textit{  after Augmentation    }}}\\
&\textbf{\textit{AP\textsubscript{50}}}  & \textbf{\textit{AP\textsubscript{s}}}  & \textbf{\textit{AP\textsubscript{m}}} & \textbf{\textit{AP\textsubscript{l}}} \\

\hline
NoAugment& 31.7 &3.8& 7.3 & 12.7 \\%& 33 & 2 & 22 & 44\\
\hline
policyV0$^{\mathrm{a}}$&  31.9(\textbf{+0.6} ) & 7.0(\textbf{+84.2} )  &7.8(\textbf{+6.8} )&11.9(\textbf{-6.2} )  \\%& 8M \\%&45/0.056 & ? \\
\hline
\textbf{policyV1}$^{\mathrm{a}}$ & 34.1(\textbf{+7.5} ) &5.9(\textbf{+55.2})& 8.2(\textbf{+12.3}) & 12.9(\textbf{+1.5})  \\%& 34 & 2 & 22 & 44\\
\hline
policyV2$^{\mathrm{a}}$&  33.0(\textbf{+4.1} ) & 6.4(\textbf{+68.4} )  &8.6(\textbf{+17.8} )&12.2(\textbf{-4} )  \\%& 8M \\%&45/0.056 & ? \\
\hline

policyV3$^{\mathrm{a}}$&  33.4(\textbf{+5.3} ) & 5.1(\textbf{+34.2} )  &7.7(\textbf{+5.4} )&13.4(\textbf{+5.5} )  \\%& 8M \\%&45/0.056 & ? \\
\hline
\multicolumn{4}{l}{$^{\mathrm{a}}$ Rotation strategies are removed from these policies.}

\end{tabular}
\label{tableaugpolicies}
\end{center}
\end{table}

\subsection{Evaluating Results}
\label{evalres}

\begin{figure*}[htbp]
\centering
  \subfloat[\label{fig4-setx1}]{\includegraphics[scale=.1]{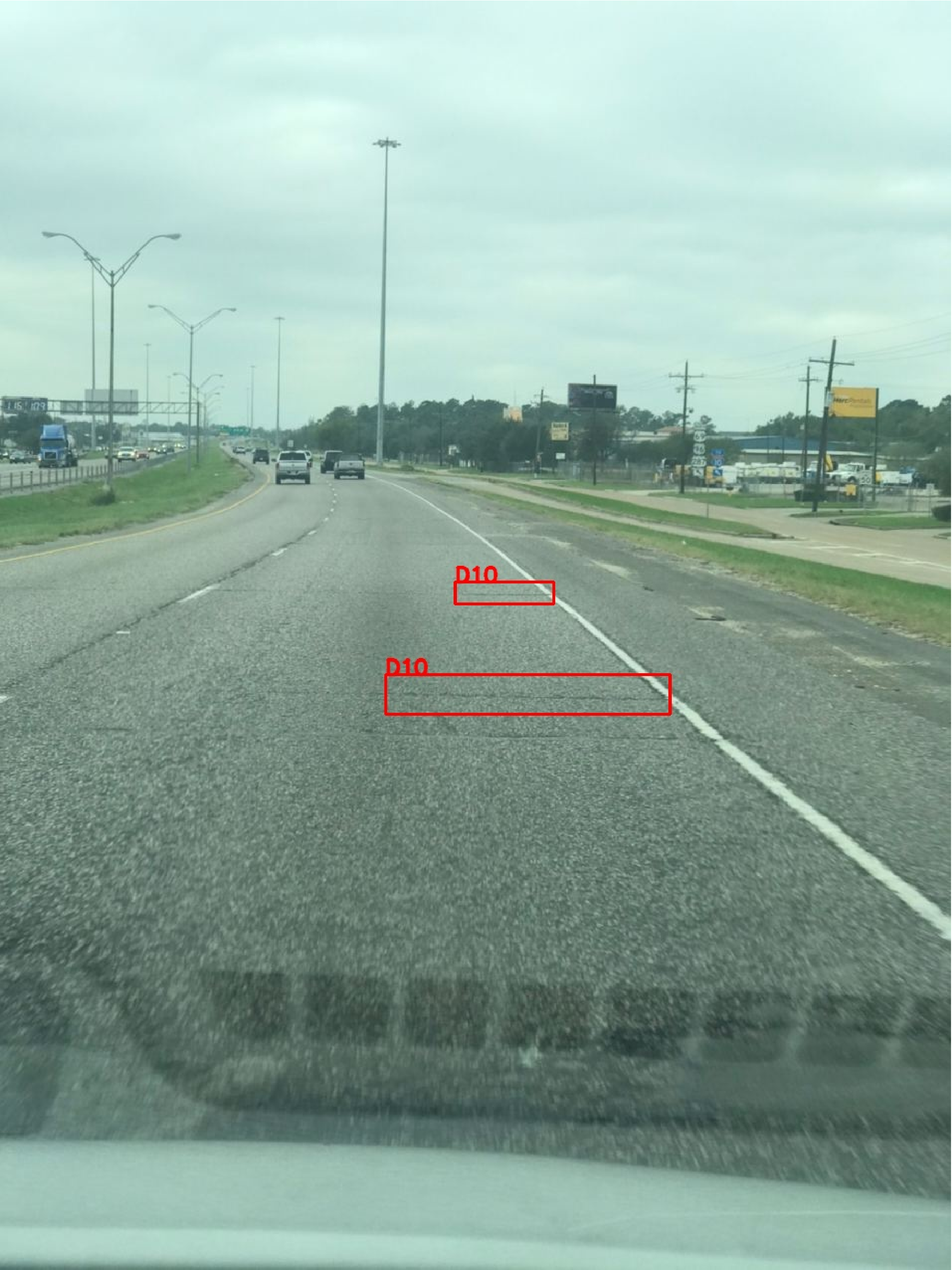}} 
  \hspace{.5cm}
  \subfloat[\label{fig4-setx2}]{\includegraphics[scale=.1] {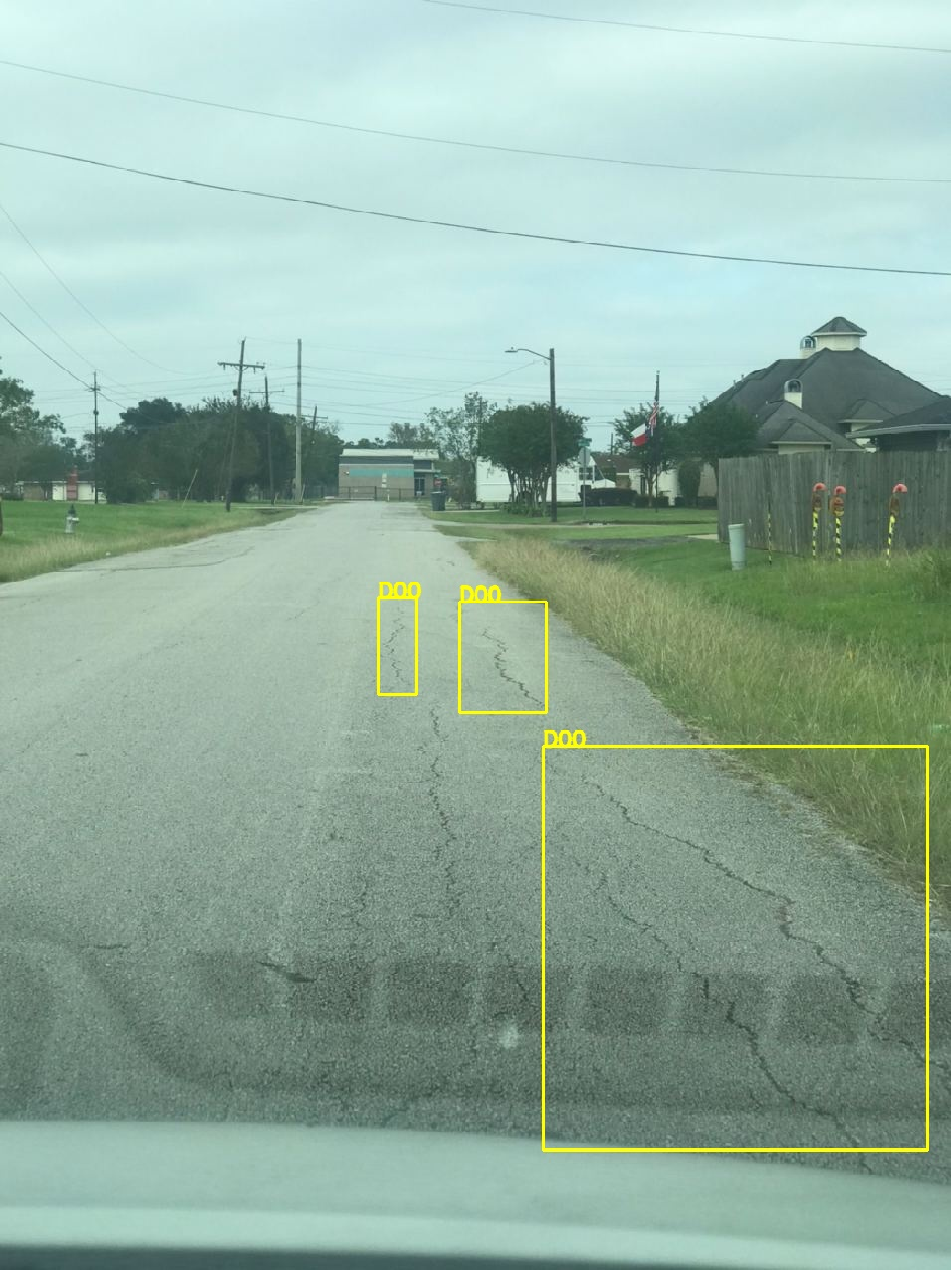}}
  \hspace{.5cm}
  \subfloat[\label{fig4-setx3}]{\includegraphics[scale=.1]{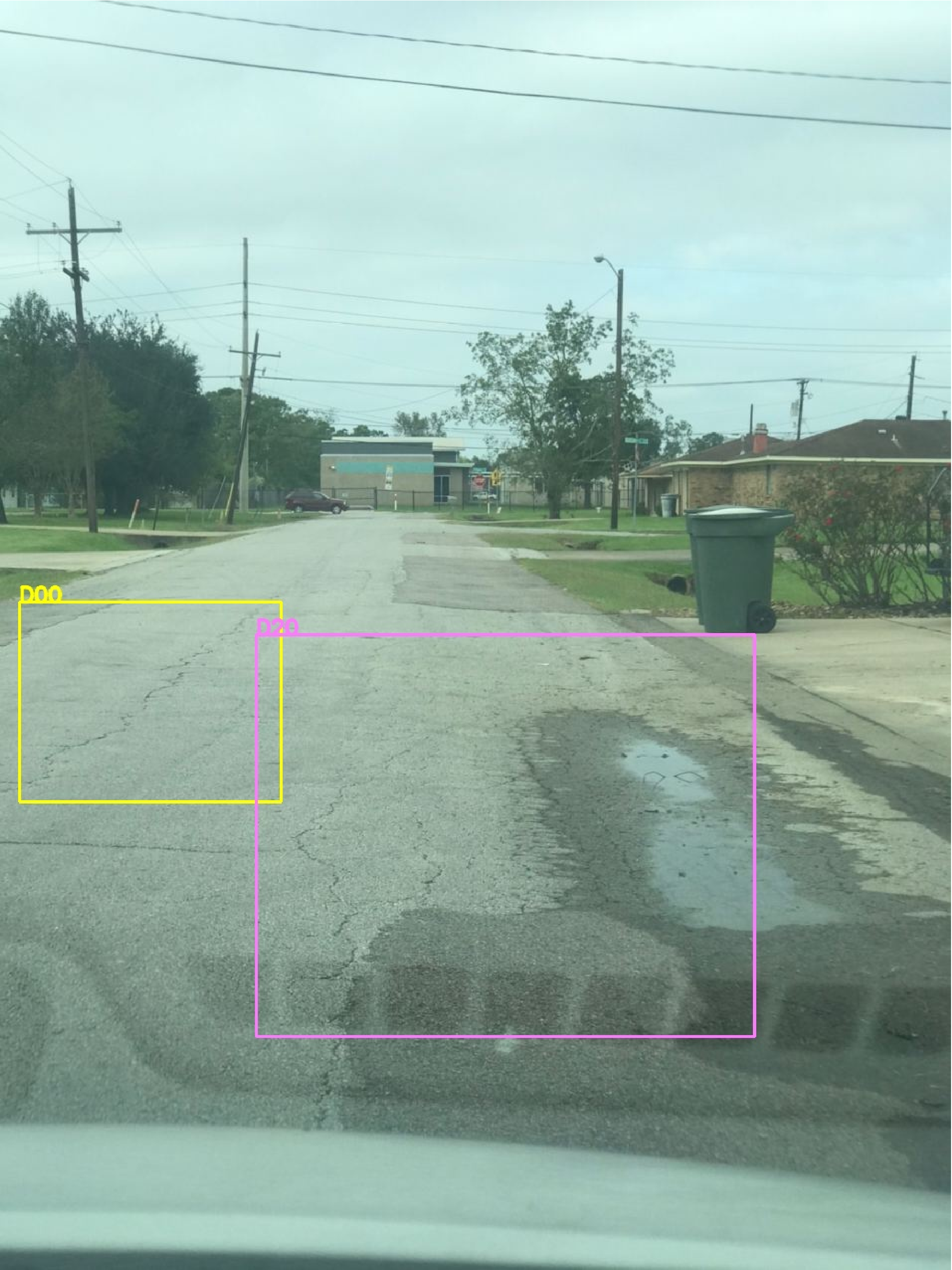}}
  \hspace{.5cm}
  \subfloat[\label{fig4-setx4}]{\includegraphics[scale=.1]{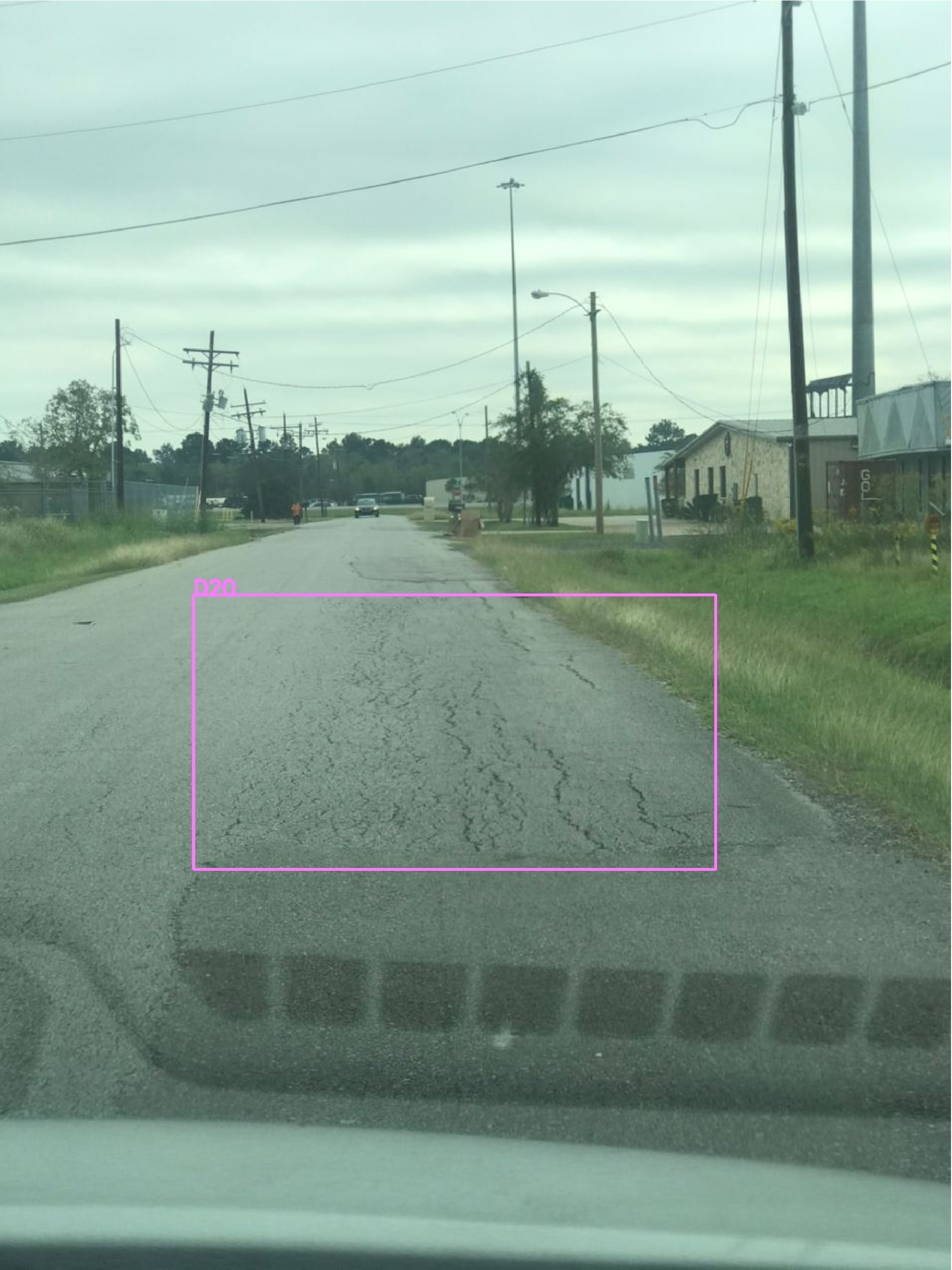}}
%   \hspace{.5cm}
%   \subfloat[\label{fig4-setx3}]{\includegraphics[scale=.1]{SETX_3.eps}}
     \caption{Examples of Network Prediction in TX: Photos are taken in South East Texas roads with smartphone showing that network is able to predict damages accurately in unseen areas.}
  \label{fig4setx} 
\end{figure*}

\begin{figure*}[htbp]
\centering
  \subfloat[\label{fig1-correct1}]{\includegraphics[scale=.25]{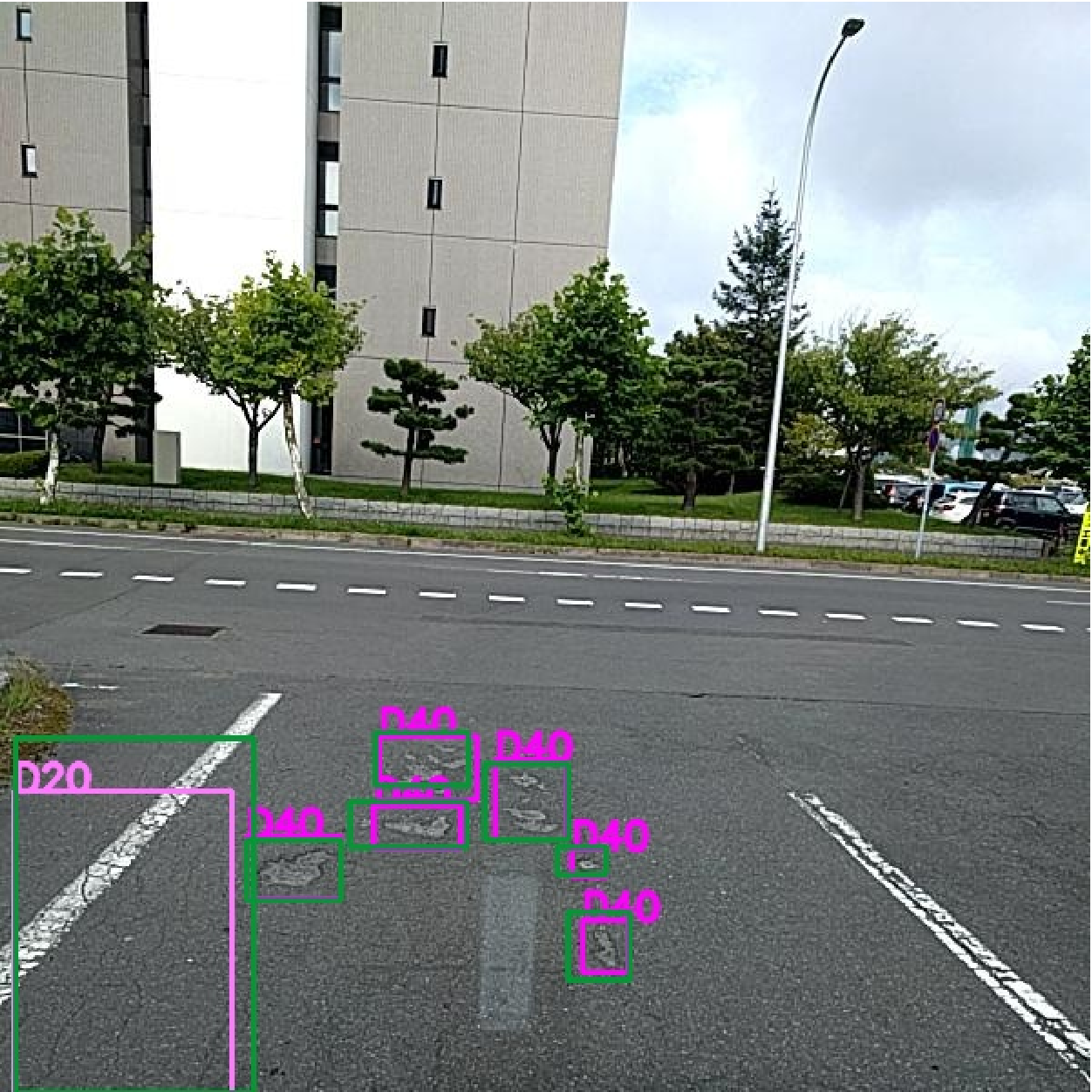}} 
  \hspace{.6cm}
  \subfloat[\label{fig1-correct2}]{\includegraphics[scale=.25] {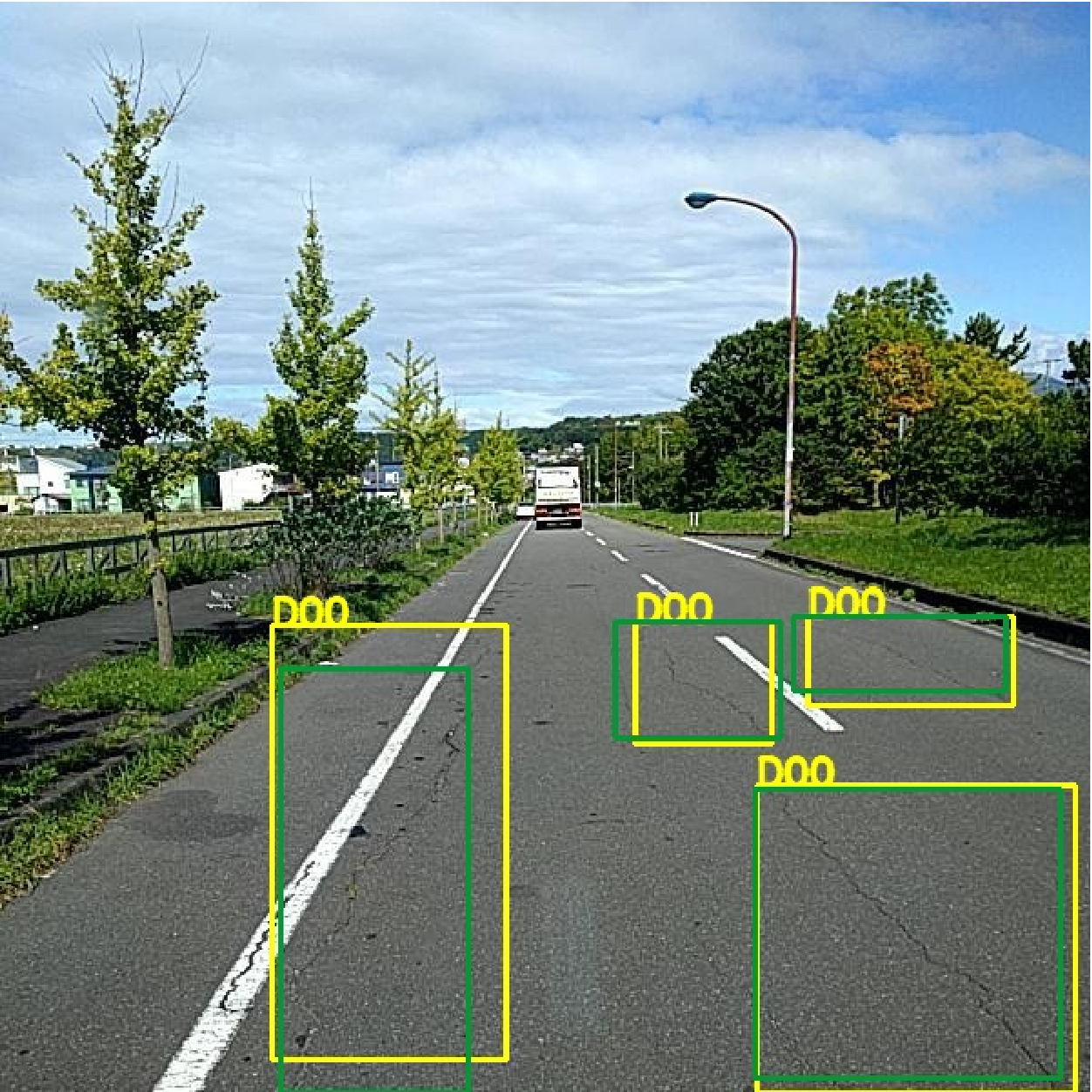}}
  \hspace{.6cm}
  \subfloat[\label{fig1-colored}]{\includegraphics[scale=.25]{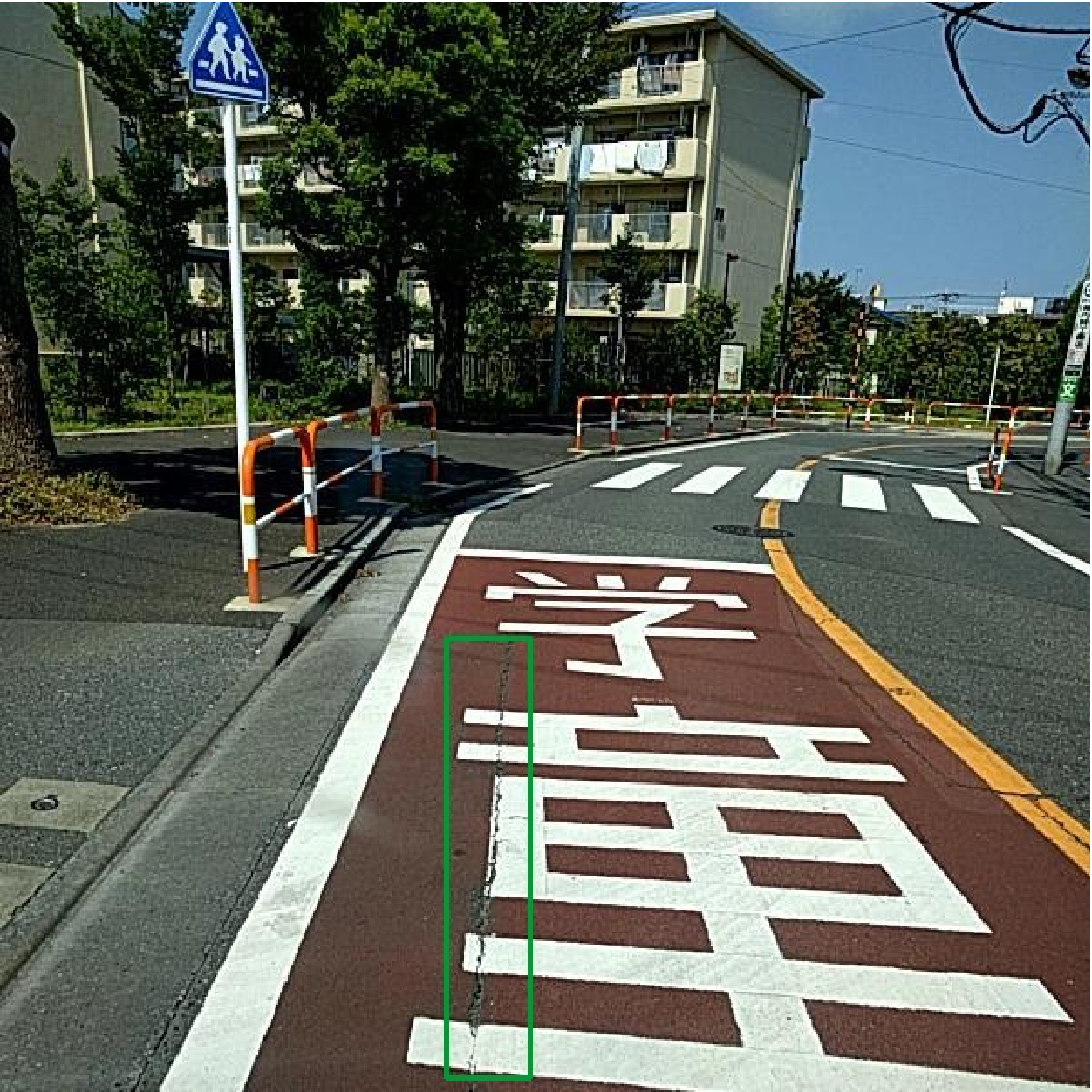}}
     \caption{Examples of Network Prediction: True Positives \ref{fig1-correct1} and \ref{fig1-correct2} (green boxes denote ground truth and others are network prediction) and  False Negatives \ref{fig1-colored}, where the background of damage is not gray. }
  \label{fig1wrong} 
\end{figure*}

\begin{figure*}[htbp]
\centering
  \subfloat[\label{fig2-jointpart}]{\includegraphics[scale=.25]{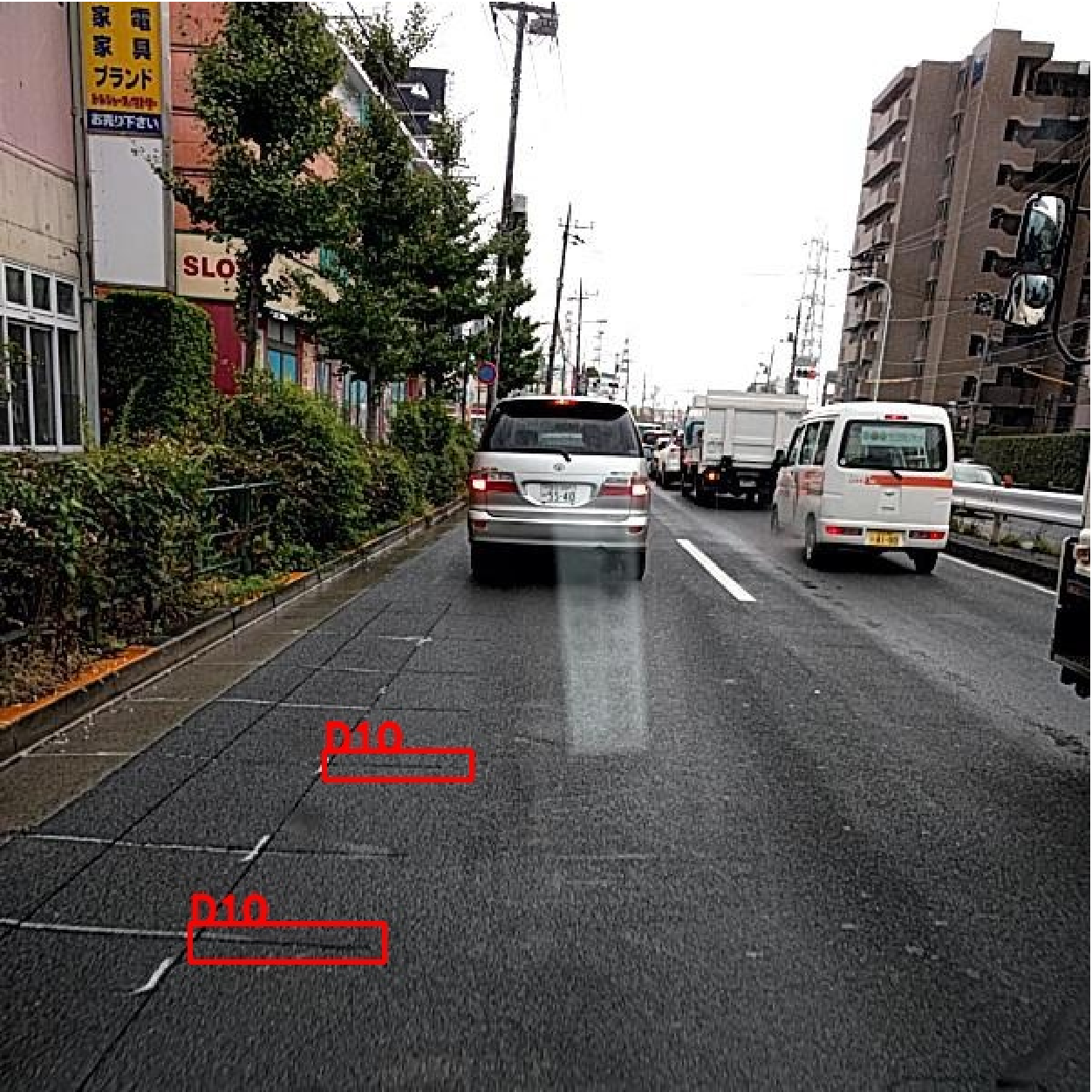}} 
  \hspace{.6cm}
  \subfloat[\label{fig2-closely}]{\includegraphics[scale=.25] {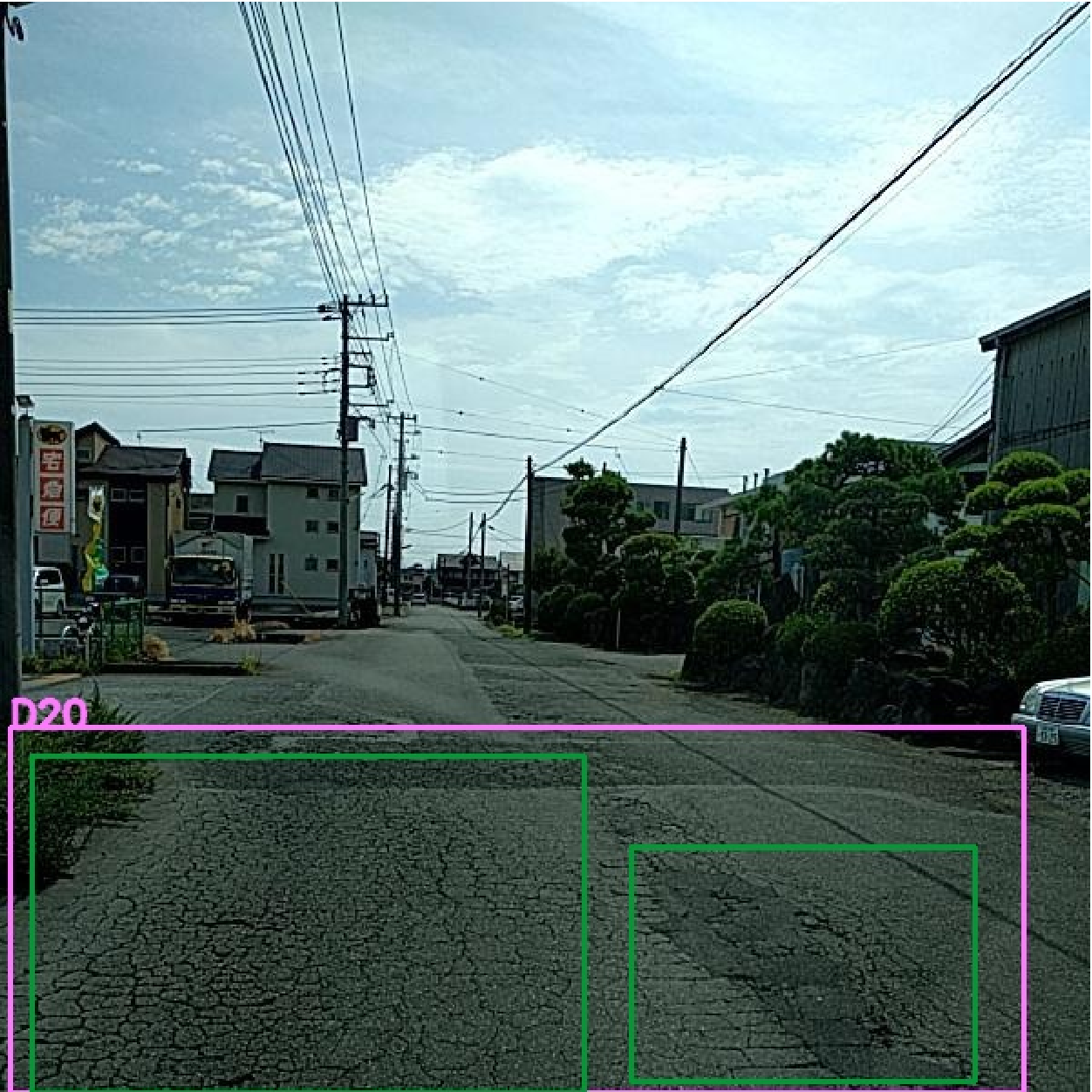}}
  \hspace{.6cm}
  \subfloat[\label{fig2-small}]{\includegraphics[scale=.25]{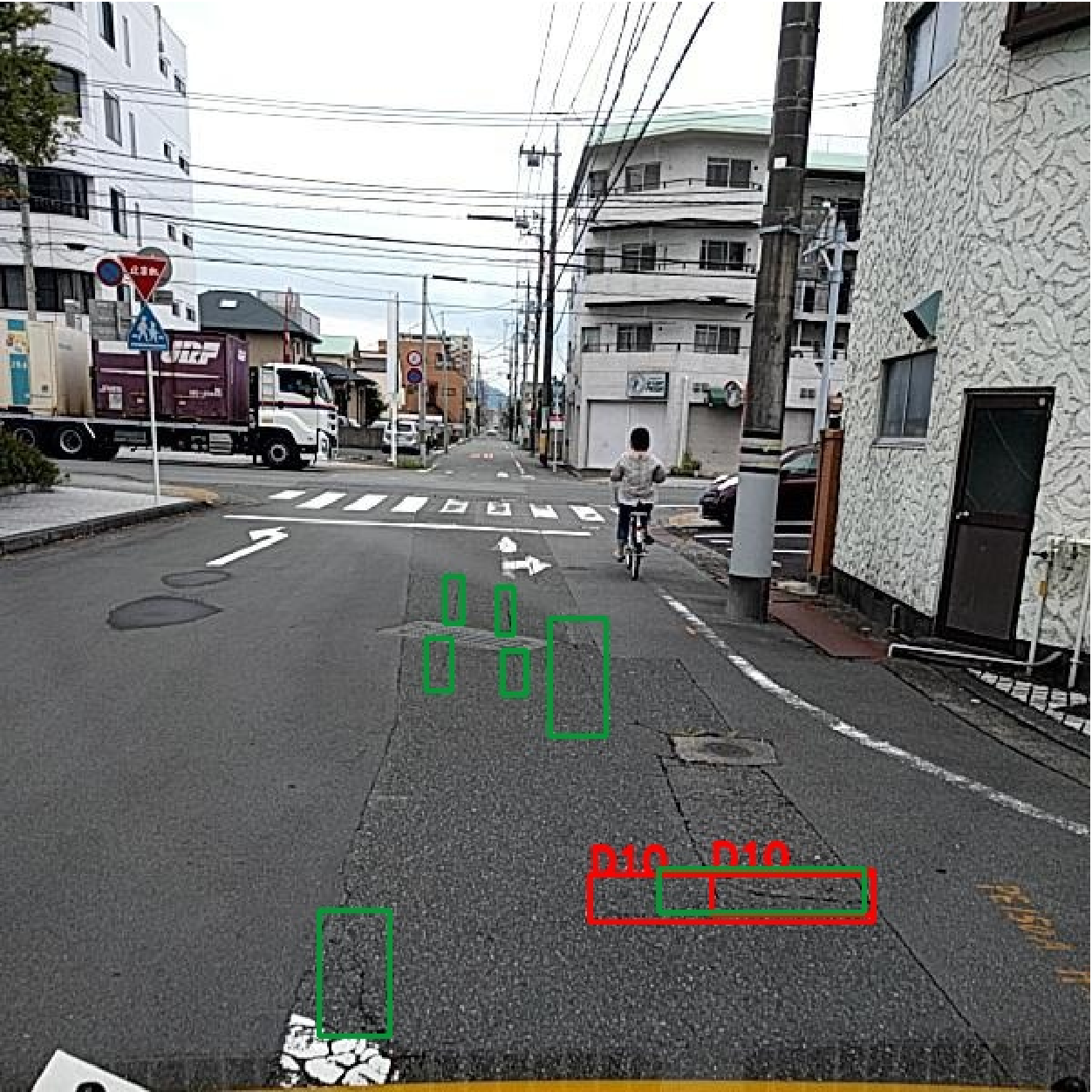}}
     \caption{Examples of Network Prediction.  False Positive of construction joint part-\ref{fig2-jointpart}, False Positive of two close bounding boxes as one crack-\ref{fig2-closely}, and False Negatives on tiny bounding boxes(D40)-\ref{fig2-small}  }
  \label{fig2wrong} 
\end{figure*}

The best F1-scores on both test sets are achieved through model D7, referring to table \ref{tabbigtable}. Some sample results from model D7 are depicted in Fig. \ref{fig4setx}, \ref{fig1wrong}, and Fig. \ref{fig2wrong}. From analyzing the results in multiple samples, it can be mentioned:
\begin{itemize}
    \item False Positive (FP) detections: There are three major contributing factors in FP detection. First, the model detects too close or overlapped bounding boxes with the same labels as a single joint bounding box, while in the ground truth the close cracks are annotated with multi bounding boxes. Although the crack is correctly detected, the IoU between the detected joint bounding box and ground truths falls behind the desired IoU threshold(i.e., 0.5). Hence, this mismatch between bounding boxes increases both FP and FN. For instance, Fig.\ref{fig2-closely} depicts two boxes with Alligator Crack(D20) annotated closely as ground truths, but the model detects the entire region as D20 type. Second, in some cases, patched surface edges(or construction joint part) are detected as a crack (commonly mistaken with Longitudinal Crack(D00) or Transverse Crack(D10) classes). As depicted in Fig. \ref{fig2-jointpart}, edges of the patched section of the pavement are detected as Transverse Crack(D10) by error. Third, manholes with area $<$ 800 pixels can be detected as potholes(D40).    
    \item False Negative(FN) detections: The contributing factors in FN detection can be counted as follows. First, a bounding box with a small area ($<$ 600), as it is shown in fig. \ref{fig2-small}. Second, different background color, e.g., a specific area or pavement markings on the road or the surface of the road is covered with debris\ref{fig1-colored}.
    \item Misclassifications: In some cases, diagonal crack misclassification has occurred between classes D00 and D10.    
\end{itemize}

Some suggested remedies for improving the results can be mentioned as follows. Based on the crack detection results, setting up some ground rules for annotating the dataset like merging close bounding boxes can decrease FP rates. Moreover, expanding the dataset with more image samples of patched pavement surfaces and manholes can help to decrease the FP rate. Also, mounting the camera in an optimal position which covers more area of the pavement would result in more practicable samples and accurate detection.

In addition, it is shown that augmentation could improve results. Referring to Table \ref{tableaugpolicies}, the results of augmentation on AP\textsubscript{50}, AP\textsubscript{s}, and AP\textsubscript{m} across all policies shows an improvement. Although the AP\textsubscript{l} results on the same table, is improved in some cases, it is observed that policies like V0 and V2 can decrease the accuracy on AP\textsubscript{l}. Many parameters could contribute to this decremental behavior, like the distribution of labels on the large boxes. Regardless of this behavior, the average improvement across all AP metrics, especially AP\textsubscript{50} and AP\textsubscript{s} can be a viable decision factor for selecting an optimal policy.    

As an instance, policies V0 and V2 are neglected because of the inclines in AP\textsubscript{l}. In contrast, policies V1 and V3 both have improved all APs. So, the policy V1 is a viable option for augmentation, because it improves AP\textsubscript{50}, AP\textsubscript{s}, and AP\textsubscript{m} further than policy V3. During training models using augmentation, it is observed that the selected policy V1 can effectively boost  AP\textsubscript{s} in smaller models D0 and D1. On the other hand, V1 for larger models, i.e., D2-D7, had less improvement over AP\textsubscript{s}. Nonetheless, utilizing augmentation does not change model F1-score for larger models. 

As the goal in this paper is set to achieve a balance between time performance and accuracy, referring to Tables  \ref{tabbigtable} and \ref{tableinf} makes it possible to choose between the D0-D7 models based on the available computational resources. For instance, for smaller embedded devices with limited power, D0 is more applicable for real-time purposes, while the accuracy is declined by 3\%. On the other hand, in the case of using a workstation with multiple GPUs, deeper models could address the real-time constraint.  Inference Time is obtained by considering the post-processing of input images. Timings with batch size 1 can be improved with further optimization of post-processing overhead.

To test the robustness and generalizability of the model, as it is mentioned in \cite{maeda2018road}, the model is tested on a set of crack images that are photographed in Southeast Texas, USA. Although the model is trained based on images from Japan, Czech, and India, it is possible to use the same model on the other regions as well. In fig. \ref{fig4setx}, four image samples from the USA are presented.

\section{Conclusions} \label{Conclusions}

In this paper, a deep-learning approach is provided to train scalable and efficient models to detect road defects, and also, the model reached competitive accuracy with 56\% F1-score. By considering Inference time, the code is able to predict in real-time using mobile devices. Although one can use TTA and ensemble learning to fuse the prediction results of the network, the inference time will dramatically increase. Moreover, transferred augmentation policies enhanced small networks by 2\% on F1-score.  %%should add more sentences

For future work, it is suggested to evaluate the reliability of the network utilizing a test set from a new country. Also to apply gradient check pointing to reduce memory cost during training and searching for optimal augmentation policies with machine learning methods and implementing bounding box the only augmentation can be studied further. Finally, to expand and clean the dataset as well as installing a camera with the optimal orientation that covers more of the pavement.

\bibliographystyle{IEEEtran}
\bibliography{IEEEabrv,main.bib}

\end{document}